\PassOptionsToPackage{unicode}{hyperref}
\PassOptionsToPackage{hyphens}{url}
\documentclass[
]{article}
\usepackage{xcolor}
\usepackage{amsmath,amssymb}
\usepackage{iftex}
\ifPDFTeX
  \usepackage[T1]{fontenc}
  \usepackage[utf8]{inputenc}
  \usepackage{textcomp} % provide euro and other symbols
  \DeclareUnicodeCharacter{2212}{\ensuremath{-}}
  \DeclareUnicodeCharacter{0394}{\ensuremath{\Delta}}
  \usepackage{dsfont}
  \newcommand{\indicator}{\mathds{1}}
\else % if luatex or xetex
  \usepackage{unicode-math} % this also loads fontspec
  \newcommand{\indicator}{\mathbb{1}}
  \defaultfontfeatures{Scale=MatchLowercase}
  \defaultfontfeatures[\rmfamily]{Ligatures=TeX,Scale=1}
\fi
\usepackage{lmodern}
\ifPDFTeX\else
\fi
\IfFileExists{upquote.sty}{\usepackage{upquote}}{}
\IfFileExists{microtype.sty}{% use microtype if available
  \usepackage[]{microtype}
  \UseMicrotypeSet[protrusion]{basicmath} % disable protrusion for tt fonts
}{}
\makeatletter
\@ifundefined{KOMAClassName}{% if non-KOMA class
  \IfFileExists{parskip.sty}{%
    \usepackage{parskip}
  }{% else
    \setlength{\parindent}{0pt}
    \setlength{\parskip}{6pt plus 2pt minus 1pt}}
}{% if KOMA class
  \KOMAoptions{parskip=half}}
\makeatother
\usepackage{longtable,booktabs,array}
\usepackage{caption}
\usepackage{calc} % for calculating minipage widths
\usepackage{etoolbox}
\makeatletter
\patchcmd\longtable{\par}{\if@noskipsec\mbox{}\fi\par}{}{}
\makeatother
\IfFileExists{footnotehyper.sty}{\usepackage{footnotehyper}}{\usepackage{footnote}}
\makesavenoteenv{longtable}
\usepackage{graphicx}
\makeatletter
\newsavebox\pandoc@box
\newcommand*\pandocbounded[1]{% scales image to fit in text height/width
  \sbox\pandoc@box{#1}%
  \Gscale@div\@tempa{\textheight}{\dimexpr\ht\pandoc@box+\dp\pandoc@box\relax}%
  \Gscale@div\@tempb{\linewidth}{\wd\pandoc@box}%
  \ifdim\@tempb\p@<\@tempa\p@\let\@tempa\@tempb\fi% select the smaller of both
  \ifdim\@tempa\p@<\p@\scalebox{\@tempa}{\usebox\pandoc@box}%
  \else\usebox{\pandoc@box}%
  \fi%
}
\def\fps@figure{htbp}
\makeatother
\providecommand{\tightlist}{%
  \setlength{\itemsep}{0pt}\setlength{\parskip}{0pt}}
\usepackage[margin=1in]{geometry}
\usepackage{bookmark}
\IfFileExists{xurl.sty}{\usepackage{xurl}}{} % add URL line breaks if available
\makeatletter
\@ifundefined{xmpquote}{}{}
\makeatother
\hypersetup{
  pdftitle={What Wins a Vote? Formatting{,} Length{,} and Lexical Diversity in the French Compar:IA LLM Arena},
  pdfauthor={Simonas Zilinskas; Maayeesha Farzana; Christophe Benavent},
  pdfsubject={Presentation associations in the French Compar:IA LLM arena},
  hidelinks,
  pdfcreator={LaTeX via pandoc}}

\title{What Wins a Vote? Formatting, Length, and Lexical Diversity in
the French Compar:IA LLM Arena}
\author{\parbox{0.96\textwidth}{\centering
Simonas Zilinskas\textsuperscript{1} \quad Maayeesha Farzana\textsuperscript{2}
\quad Christophe Benavent\textsuperscript{2}\\[0.45em]
{\small\textsuperscript{1}Compar:IA \quad
\textsuperscript{2}Université Paris Dauphine-PSL}\\[0.45em]
\href{mailto:contact@comparia.beta.gouv.fr}{\texttt{contact@comparia.beta.gouv.fr}}}}
\date{Preprint v1.0, 30 September 2026}

\begin{document}
\maketitle

\begin{abstract}

LLM arenas turn pairwise human preferences into model rankings. Those
preferences may reflect how an answer is presented as well as what it
says. We take a stylometric approach to 137,293 decisive
French-language votes from the July 2026 Compar:IA release; the primary formatting analysis includes 137,113
battles across 116 models, and the joint estimates use the 127,092
battles with all required measurements. For each battle, we reconstruct
the response visible when the user voted. We then compare the raw
ranking with rankings adjusted for formatting, length, readability,
vocabulary variety, and sentence structure. Presentation is associated
with winning, but length, bold text, and lists tend to occur together,
making their individual contributions hard to separate. Across the
measured features, two associations change least across specifications:
\textbf{bold usage} (+11.0\% win odds per standard deviation in the joint
model) and \textbf{moving-average type-token ratio (MATTR)}, a measure of
vocabulary variety that is less sensitive to answer length (+16.8\%).
The bold association is substantially smaller in observed multi-turn
conversations, whereas the MATTR association changes little; because
users choose whether to continue, this difference is descriptive rather
than causal. The full adjustment moves 36 of 116 models by at least ten
ranks. Yet comparisons with external benchmarks do not show that
adjusted rankings better measure capability. We therefore recommend
publishing raw and adjusted rankings side by side as a transparent
sensitivity analysis.
\end{abstract}


\section{Introduction}\label{introduction}

LLM evaluation arenas show users two anonymous responses and ask which
they prefer. The resulting rankings measure human preference. They do
not directly measure factual accuracy or technical capability.
Compar:IA, a French government-backed arena launched publicly in October
2024, describes its ranking in the same terms (Compar:IA, 2026a).

This distinction matters because users may respond to presentation as
well as substance. Previous studies have found preferences related to
response position, length, verbosity, and errors, although the pattern
depends on whether the evaluator is a person, an LLM judge, or a reward
model (Zheng et al., 2023; Singhal et al., 2024; Saito et al., 2023;
Dubois et al., 2024; Wu \& Aji, 2025). Li et al.~(2024) addressed this
problem in LMArena by adding response length and markdown features to
its ranking model. The adjusted ranking changed, but the study could not
determine whether style was a bias or a genuine signal of answer
quality.

We extend that analysis beyond markdown and length, and we do so
stylometrically: we describe each answer by measurable properties of its
surface form and ask which of those properties track the vote. We study
three groups of observed response properties: \textbf{formatting} (bold,
lists, headers, code, and emoji), \textbf{length}, and \textbf{language}
(readability, vocabulary variety, and sentence structure). These
properties are correlated: longer answers, for example, tend to
contain more lists and bold text. We estimate them jointly to ask which
associations remain after accounting for the other measured properties
and for the model that produced each response.

These properties also become available to a voter in different ways.
Bold text, headings, and code blocks are visually apparent, while
vocabulary variety emerges only through reading. Compar:IA records
whether a vote followed one user turn or several, which lets us ask
whether these associations differ across conversation contexts. Turn
count does not tell us how carefully anyone read, however, and users
decide for themselves whether to continue a conversation.

Compar:IA offers about 137,000 decisive French-language votes across 116
models. It also records the turn at which the vote occurred, which
allows us to exclude later conversation turns from every measurement.
Its broad model roster lets us distinguish patterns that arise because
different models have different presentation habits from patterns
observed across answers by the same model.

We ask three questions: \textbf{Which aspects of presentation remain
associated with votes after the other measured properties are taken
into account? Do those associations differ between single-turn and
multi-turn conversations? How much do the adjustments change the
ranking?} Our analysis makes four contributions. First, it estimates
formatting, length, and language in one ranking model and shows that many
of their associations overlap. Second, it identifies bold usage and
comparatively length-robust vocabulary variety as the two most stable
correlates. Third, it reconstructs every response as it appeared at the
time of the vote and tests how the associations vary with observed
conversation depth. Fourth, it compares the resulting rankings with
external leaderboards to test, rather than assume, that adjustment
brings arena rankings closer to independent capability measures.

\section{Data}\label{data}

\subsection{The Compar:IA Platform}\label{the-comparia-platform}

Compar:IA is an LLM evaluation arena operated by the French government's
Ministry of Culture and Direction interministérielle du numérique
(DINUM). Users submit prompts and receive responses from two anonymous
models side-by-side, may continue the conversation over several turns,
and then vote for a winner or declare a tie. Models are identified only
after voting.

The platform offers several arena modes; in our data the decisive votes
come from \textbf{random} (72\%, random model pairs), \textbf{custom}
(19\%, user-selected pairs), \textbf{big-vs-small} (8\%, deliberately
pairing large and small models), and \textbf{small-models} (2\%).

\subsection{Dataset}\label{dataset}

We use \textbf{\texttt{ministere-culture/comparia-fr-arena}} (Compar:IA
\& Ministère de la Culture, 2026b), the consolidated Compar:IA release
published on Hugging Face under Open Licence 2.0 (Etalab) and CC-BY-4.0.
We pin revision
\textbf{\texttt{8cd6488c5d0c3b8dfcb9339d11ae9624c84359be}} (published
and accessed 8 July 2026). The gated release contains 641,277 turn rows
and roughly 208K human reactions across at least 119 models before our
French/decisive filtering. It is organised by turn: one row per
conversation turn, and \texttt{choice} records any reaction made on that
turn. A small number of conversations contain multiple decisive
reactions; we retain the last decisive reaction per
\texttt{comparison\_id}, yielding one battle per conversation.

{\def\LTcaptype{none} % do not increment counter
\begin{longtable}[]{@{}lrrr@{}}
\toprule\noalign{}
& Raw turns & Decisive French battles & Models (\(\geq\)100 battles) \\
\midrule\noalign{}
\endhead
\bottomrule\noalign{}
\endlastfoot
comparia-fr-arena & 641,277 & 137,293 & 116 \\
\end{longtable}
}

Ties and no-vote turns are dropped (decisive votes only, matching prior
style-control work). Winner distribution is balanced (model A wins
49.9\%, see §4.1).

\subsection{Vote-Time Reconstruction, Features, and Data
Quality}\label{feature-extraction-and-data-quality}

For each battle, we reconstruct only the text visible when the vote was
cast. We concatenate the assistant messages up to that point and sum the
release's per-turn token counts over the same period. We do not use
token totals for the completed conversation. We then compute formatting
and language features on this vote-time text. Battles with no visible
response text at the vote are dropped.

The release can store hidden reasoning separately in
\texttt{reasoning\_content} or embed it in a paired
\texttt{\textless think\textgreater...\textless/think\textgreater} span
before the final answer. We never analyse \texttt{reasoning\_content}.
For embedded spans, we remove each complete span and retain only text
outside it; if a tag is unmatched, we retain only the unambiguous text
before it. Compared with a stricter rule that drops any response
containing a tag, this rule keeps 86 more battles with visible final text
and drops 7 ambiguous cases, a net increase of 79. A versioned aggregate audit
(\texttt{results/reasoning\_content\_audit\_results.json}) of the pinned
release found no retained decisive French vote prefix whose last
assistant message had empty final \texttt{content} but non-empty
\texttt{reasoning\_content}. We therefore find no evidence of that directly observable
missing-final-answer failure mode in the pinned release. An untagged
reasoning trace stored in \texttt{content} cannot be identified from
these fields alone, so it cannot be ruled out.

Topic comes from the release metadata, and conversation depth is the
number of user turns visible at the vote. Topic is present for 100\% of
battles and length for 100\%. Readability is available
for 97.2\%, but all language features are jointly available for 92.7\%,
which is why the complete-case model uses fewer battles.

\paragraph{Topic metadata and vote-timing audit.}\label{native-topic-metadata-and-vote-timing-audit}

Each conversation has an LLM-assigned topic from about 18 subject
classes, which we use in §4.4. It also records when each reaction
occurred. Some source conversations continued after the retained vote:
this happened in \textbf{15,357 of 137,293 battles (11.2\%)}. Those
later turns are excluded from the response text, token counts, and depth
measures. At the time of voting, 15,552 battles were genuinely
multi-turn. An independent audit against the raw release found that no
post-vote turn entered any measured feature.

\section{Methodology}\label{methodology}

\subsection{Presentation Features}\label{presentation-features}

We measure presentation in three families, all computed per response.

\textbf{Formatting}, five markdown features:

{\def\LTcaptype{none} % do not increment counter
\begin{longtable}[]{@{}
  >{\raggedright\arraybackslash}p{(\linewidth - 4\tabcolsep) * \real{0.2432}}
  >{\raggedright\arraybackslash}p{(\linewidth - 4\tabcolsep) * \real{0.3514}}
  >{\raggedright\arraybackslash}p{(\linewidth - 4\tabcolsep) * \real{0.4054}}@{}}
\toprule\noalign{}
\begin{minipage}[b]{\linewidth}\raggedright
Feature
\end{minipage} & \begin{minipage}[b]{\linewidth}\raggedright
Description
\end{minipage} & \begin{minipage}[b]{\linewidth}\raggedright
Regex Pattern
\end{minipage} \\
\midrule\noalign{}
\endhead
\bottomrule\noalign{}
\endlastfoot
\textbf{Headers} & Markdown headers (\# through \#\#\#\#\#\#) &
\texttt{\^{}\#\{1,6\}\textbackslash{}s} (multiline) \\
\textbf{Lists} & Ordered and unordered list items &
\texttt{\^{}\textbackslash{}s*{[}-*+{]}\textbackslash{}s} and
\texttt{\^{}\textbackslash{}s*\textbackslash{}d+\textbackslash{}.\textbackslash{}s} \\
\textbf{Bold} & Bold-formatted text &
\texttt{\textbackslash{}*\textbackslash{}*{[}\^{}*{]}+\textbackslash{}*\textbackslash{}*} \\
\textbf{Code blocks} & Fenced code blocks &
\texttt{\textasciigrave{}\textasciigrave{}\textasciigrave{}} or
\texttt{\textasciitilde{}\textasciitilde{}\textasciitilde{}} (counted in
pairs) \\
\textbf{Emoji} & Emoji characters & Unicode emoji ranges \\
\end{longtable}
}

\textbf{Length}, the response's cumulative output-token count from the
dataset metadata.

\textbf{Language}, covering properties beyond formatting: readability
(the Kandel-Moles reading ease, REL, calibrated for French; Coleman-Liau; Flesch-Kincaid
grade), lexical diversity, and sentence structure (Kandel \& Moles,
1958; Coleman \& Liau, 1975; Kincaid et al., 1975). We use two diversity
measures. Type-token ratio (TTR) is the number of unique words divided
by the total number of words, so it tends to fall as answers get longer.
Moving-average type-token ratio (MATTR) computes the same ratio within
a rolling window and is less dependent on total length (Covington \&
McFall, 2010); we use a 50-token window.

Section 4.1 first reproduces a narrow formatting analysis that is
structurally similar to prior style control. Section 4.2 then adds
length and all language features. We separate these specifications
because length may reflect genuine completeness as well as presentation.
Controlling for it can therefore remove useful quality signal, not just
bias. Our coefficients are not numerically comparable with Li et
al.~(2024), who standardise a relative feature difference, while we
standardise the raw difference between responses A and B.

\subsection{Bradley-Terry Model and Statistical Inference}\label{bradley-terry-model}

We use a Bradley-Terry (BT) model, a standard method for turning
pairwise wins and losses into model ratings (Bradley \& Terry, 1952).
The basic model asks how strongly each model's identity predicts a win.
The controlled model also compares the presentation of the two
responses. It therefore estimates how the ranking changes after
accounting for measured presentation differences.

\textbf{Standard model.} For each decisive battle, the outcome is 1 when
model A wins and 0 when model B wins. Model A receives +1 and model B
receives −1 in the corresponding model columns. We fit logistic
regression without regularisation. The all-model contrast design is
identified only up to an additive constant, so after fitting we recentre
the model coefficients to sum to zero. This makes the mean rating 1000
without changing any pairwise prediction:

\[\text{Rating}_i = 1000 + \frac{400 \cdot \beta_i}{\ln(10)}\]

The value 1000 is only a reference point. The factor 400 follows the
usual Elo-style convention: a 400-point rating gap corresponds to a
tenfold difference in predicted win odds. It changes the reporting
scale, not the fitted probabilities or the ordering of models.

\textbf{Style-controlled model.} Following the style control of Li et
al. (2024), which adapts the regression-based length control of Dubois
et al. (2024) to the Bradley-Terry model, we add the difference between
responses A and B for each presentation feature. These differences are
standardised, so a coefficient represents the association with a
one-standard-deviation difference:

\[P(\text{A wins}) = \sigma\left(\sum_i \beta_i \cdot \indicator_i + \sum_f \gamma_f \cdot \Delta f\right)\]

Here, \(\sigma\) is the logistic function that converts the linear score
into a probability. The indicator \(\indicator_i\) is +1 for model A,
−1 for model B, and 0 for models not in the battle. For feature \(f\),
\(\Delta f\) is the standardised difference between responses A and B,
and \(\gamma_f\) is its estimated association with the vote. The
coefficient \(\beta_i\) represents the remaining preference for model \(i\)
after measured presentation is included; it should not be read as pure
skill. The \(\gamma_f\) terms describe the remaining association between
each measured feature and the vote. Including a coefficient for every
model means that these feature associations are estimated after
accounting for stable differences between models. The primary
formatting-only analysis in §4.1 uses standardised raw contrasts.
The joint, conversation-depth, and topic models (both the within-topic
fits and the topic-interaction model) cap each feature contrast at the 1st and 99th percentiles before standardising so
that a few unusual battles do not dominate those estimates. The
exploratory task-stratified fits follow the raw-contrast formatting
specification.

\paragraph{Uncertainty and robustness.}\label{bootstrap-inference}

Primary coefficient and model-rating intervals use 1,000 nonparametric
bootstrap samples of battles drawn with replacement (Efron, 1979). The
exploratory topic and task strata use 400 samples, the weekly block
sensitivity analysis uses 500 samples, and external paired-correlation
comparisons use 10,000. Battle-level resampling treats battles as
independent and cannot account for repeated users or other latent
clustering. Weekly resampling is a temporal sensitivity analysis, not a
substitute for user-level clustering.

\paragraph{Multiple comparisons.}\label{multiple-comparison-correction}

Because we test several features and model-rating changes, we apply the
Benjamini-Hochberg procedure at 0.05 to limit false discoveries
(Benjamini \& Hochberg, 1995). We correct the five primary formatting
features, the thirteen joint-model features, the ten formatting
interactions, each conversation-depth interaction family, and per-model
rating changes as separate families.
Rank changes are descriptive and are not tested separately. Topic- and
task-stratified intervals are unadjusted exploratory intervals and are
described as such.

\section{Results}\label{results}

We begin with formatting and its effect on the ranking, then add length
and language. We next examine conversation depth, subject and task, and
the main robustness checks. We end by comparing the adjusted rankings
with production and external benchmarks.

\subsection{Formatting Associations and Ranking Sensitivity}\label{formatting-and-votes}

Bold has the largest formatting association with votes. Table 1 reports
each feature twice: fitted alone, and fitted together with the other
four, in both cases after accounting for model identity.

\textbf{Table 1. Formatting coefficients from the Bradley-Terry model
(137,113 battles, 116 models), each feature fitted alone and all five
fitted together. Odds change per standard deviation of the A-minus-B
contrast. Intervals, p-values, and significance refer to the joint fit;
p-values are BH-adjusted across the 5 tests.}

{\def\LTcaptype{none} % do not increment counter
\begin{longtable}[]{@{}lrrlll@{}}
\toprule\noalign{}
Feature & Alone & Together & 95\% CI & p (BH) & Significant? \\
\midrule\noalign{}
\endhead
\bottomrule\noalign{}
\endlastfoot
\textbf{Bold} & +27.0\% & \textbf{+19.0\%} & {[}+16.0\%, +22.4\%{]} &
0.002 & Yes \\
\textbf{Headers} & +22.3\% & +9.1\% & {[}+3.0\%, +12.6\%{]} & 0.002 &
Yes \\
\textbf{Lists} & +16.8\% & +6.6\% & {[}+4.5\%, +9.0\%{]} & 0.002 &
Yes \\
\textbf{Emoji} & +8.8\% & +3.5\% & {[}+1.5\%, +5.4\%{]} & 0.002 &
Yes \\
Code blocks & +4.8\% & +0.8\% & {[}−0.8\%, +4.5\%{]} & 0.580 & No \\
\end{longtable}
}

\textbf{Interpretation.} A one-standard-deviation difference in bold
usage is associated with 19.0\% higher win odds, after adjusting for
model identity and the other formatting features. Headers, lists, and
emoji have smaller but significant associations; code blocks show no
detectable pooled association. Bold is the strongest formatting
correlate in this specification.

\paragraph{Formatting features share the same signal.}\label{formatting-features-share-the-same-signal}

Every estimate shrinks when the other four are added, in some cases
sharply: headers and lists each fall by about 60\%. A feature tested
alone is credited with the pattern it shares with the rest. The data
therefore cannot cleanly separate features that usually appear together,
and neither column of Table 1 should be read as that feature's own
contribution.

\paragraph{No evidence of formatting interactions.}\label{formatting-interactions}

An additive model would be the wrong choice if formatting worked in
combinations, so that bold mattered more in an answer already broken
into headed sections. We tested this by adding all ten pairwise products
of the standardised formatting contrasts to the joint fit. None of the
ten survives BH correction across that family. The largest are headers ×
code blocks (−7.1\%, 95\% CI {[}−13.1, −0.4{]}) and bold × code blocks
(+3.8\% {[}+0.5, +7.7{]}); bold × headers is −3.8\% {[}−9.6, +1.1{]}.
The five main effects move by at most one point (bold +19.3\%, headers
+9.9\%, lists +7.0\%, emoji +3.7\%, code blocks +0.5\%), and in-sample
classification accuracy does not improve. A likelihood-ratio test is
significant (\(\chi^2\) = 60.6 on 10 degrees of freedom, p \(<\) 0.001),
but with 137,113 battles even very small effects reach significance, and
the extra terms do not improve classification. Because several formatting features are sparse
and strongly correlated, we report this as a specification check rather
than as evidence about particular bundles, and we keep the additive
model for the rest of the paper.

\paragraph{How much the ranking changes.}\label{how-much-the-ranking-changes}

The Pearson correlation between standard and formatting-controlled BT ratings is
\textbf{r = 0.985}: rankings are stable overall, but specific models
shift. Across all 116 models, \textbf{87 (75\%)} show a statistically
significant rating change after style control.

\textbf{Table 2. Largest rank changes after formatting control. ``Sig?''
refers to the BH-adjusted test of the underlying rating change.}

{\def\LTcaptype{none} % do not increment counter
\begin{longtable}[]{@{}lcrl@{}}
\toprule\noalign{}
Model & Std → Ctrl & ΔRank & Sig? \\
\midrule\noalign{}
\endhead
\bottomrule\noalign{}
\endlastfoot
mistral-small-2603 & 9 → 33 & \textbf{−24} & Yes \\
qwen-3-8b & 80 → 102 & \textbf{−22} & Yes \\
mistral-large-2512 & 3 → 24 & \textbf{−21} & Yes \\
gpt-oss-120b & 37 → 56 & \textbf{−19} & Yes \\
gpt-5.3 & 45 → 27 & \textbf{+18} & Yes \\
qwen3-30b-a3b & 78 → 95 & \textbf{−17} & Yes \\
qwen3-32b & 54 → 68 & \textbf{−14} & Yes \\
mistral-medium-2508 & 2 → 15 & \textbf{−13} & Yes \\
o4-mini & 76 → 63 & \textbf{+13} & Yes \\
\end{longtable}
}

Heavy formatters (mistral-large-2512, mistral-small-2603, gpt-oss-120b)
fall, while several more concise models rise. These shifts establish
sensitivity to the adjustment; they do not by themselves show that the
models which rise have greater reasoning depth or substantive quality.

\paragraph{Position bias and arena modes.}\label{position-bias-and-arena-modes}

Position bias is negligible and not significant: model A wins 49.94\% of
decisive battles (binomial p = 0.67). Win rates are close to 50\% within
every arena mode (random 49.9\%, custom 50.1\%, big-vs-small 49.6\%,
small-models 50.7\%), consistent with the balanced construction of the
consolidated release.

\subsection{Adding Length and Language}\label{adding-length-and-language}

Adding length and language features changes the picture: much of the
formatting association is shared with answer length, while bold and
MATTR remain distinct signals. The retained frame has 137,293 battles.
The primary formatting analysis excludes 180 battles involving three
models below the 100-battle threshold, leaving 137,113. For the joint
analysis, missing readability removes 5,891 battles, missing TTR removes
another 187, and missing MATTR removes another 3,950; applying the model
threshold then removes 173, leaving \textbf{127,092 battles across 116
models}. Length and formatting are complete in the retained frame.

\textbf{Some of the formatting association is also a length
association.} Longer answers tend to contain more markdown, so adding
length makes the formatting estimates smaller. Length itself is
associated with +8.3\% win odds {[}+5.7, +10.6{]}. This does not mean
that length is merely a bias, since a longer answer may also be more
complete.

\textbf{Bold remains associated with votes after all measured features
are included.} On the joint model's common support, its formatting-only
estimate is +20.4\%; after length and language are added it is
\textbf{+11.0\%} {[}+8.8, +13.6{]}. This same-support comparison
separates specification change from sample change. Headers are +6.8\%,
code blocks +2.4\%, emoji +3.3\%,
and lists +1.7\%. Lists are not significant after correction; the other
four formatting estimates are. The small list estimate also changes
across specifications.

\textbf{MATTR is the most stable language measure.} A
one-standard-deviation increase is associated with \textbf{+16.8\%} win
odds {[}+14.9, +18.9{]}. MATTR is nearly unrelated to total length in
these data, so it captures something different from simply writing more.
Raw TTR is −26.3\%, but TTR mechanically falls as text gets longer and
should not be interpreted as an independent diversity signal. Taken
together, length, markdown density, and raw TTR overlap too much for
their individual estimates to have a simple interpretation. MATTR stands
apart from that group.

\textbf{Readability and sentence length add little.} REL (−0.8\%) and
Flesch-Kincaid (−3.7\%) are not significant. Coleman-Liau is +3.4\%, but
the three readability measures overlap strongly, so we do not interpret
one as uniquely important. Mean sentence length is not significant. The
share of long sentences is significant but small (−3.6\%
{[}−5.3, −2.0{]}), and is the only sentence-structure signal the joint
model retains.

\begin{figure}[htbp]
\centering
\pandocbounded{\includegraphics[width=\textwidth,keepaspectratio,alt={Figure 1. Joint presentation coefficients and nested formatting specifications. The left panel reports joint-model log-odds coefficients with 95 percent bootstrap intervals; the right panel shows how formatting coefficients change after adding length and linguistic features.}]{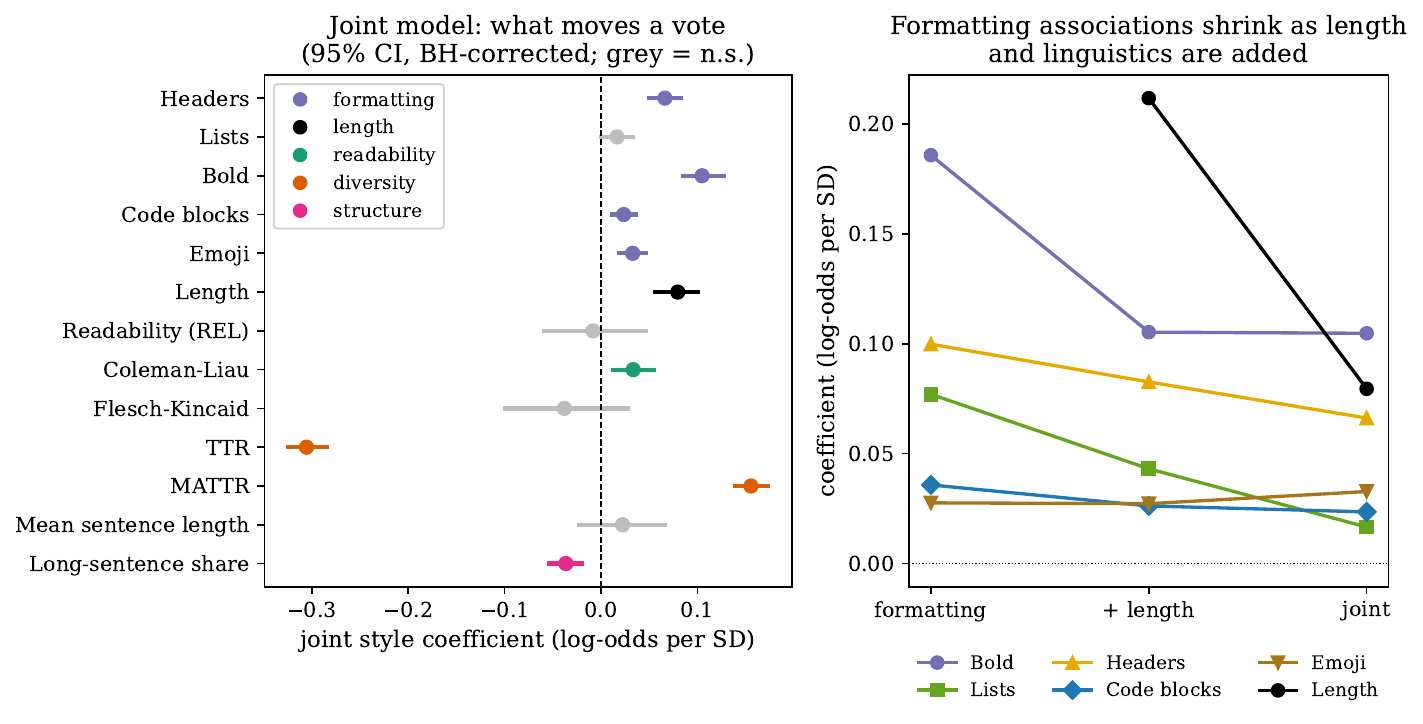}}
\caption{Left: joint style coefficients with 95\% bootstrap
CIs, coloured by feature family (grey = not significant after BH).
Bold and MATTR remain clearly positive; raw TTR is strongly negative
because it falls mechanically with length. Right: formatting
coefficients shrink as length and then the linguistic features are
added. Length has the largest coefficient when added to formatting and
shrinks substantially once the linguistic features, which overlap with
it, are included.}
\label{fig:joint-coefficients}
\end{figure}

\textbf{The added features provide a small gain in in-sample fit.}
In-sample battle-classification accuracy rises from 0.635 with
formatting alone to 0.637 after adding length and \textbf{0.642} in the
full model. This descriptive improvement does not measure out-of-sample
performance and does not show that presentation caused the votes.

\textbf{The joint control reshuffles the leaderboard more than
formatting alone.} On the common support, the standard ratings
correlate with the formatting-only controlled ratings at Pearson
r = 0.967 but with the \emph{joint} controlled ratings at only
\textbf{0.931}; on ranks the joint comparison is Spearman 0.915. \textbf{36 of 116 models move by \(\geq\)10 ranks}. Heavy
formatters fall (gpt-oss-120b −48, mistral-small-2603 −35,
mistral-large-2512 −33, glm-4.5 −29), while several more concise models rise
(gpt-5.3 +43, claude-3-5-sonnet-v2 +34, trinity-large-preview +29,
claude-3-7-sonnet +28, o4-mini +28).

\textbf{Much of the raw pattern comes from differences between models.}
Without model controls, the estimates are larger: bold is +29.1\%, mean
sentence length +23.7\%, and MATTR +32.3\%. After accounting for which
model produced each answer, they fall to +11.0\%, +2.3\%, and +16.8\%.
Models that usually present answers differently therefore explain a
substantial part of the pooled pattern. The remaining association is
estimated after stable model differences are accounted for, but it is
still not causal and should not be read as a within-model causal effect.

\subsection{Single-Turn and Multi-Turn
Conversations}\label{single-turn-and-multi-turn-conversations}
Formatting is less strongly associated with votes in multi-turn
conversations, especially for bold. We compare battles with one user
turn visible at the vote against those with two or more. Of 137,113
battles in this analysis, 88.7\% are single-turn and 11.3\% are
genuinely multi-turn at the time of voting.

We estimate both groups in one model by allowing each presentation
association to change for multi-turn battles. A negative interaction in
Table 3 means that the association is smaller in multi-turn
conversations. Bold, headers, and code blocks have significant negative
interactions after correction; lists does not. Emoji has a small
positive interaction.

\textbf{Table 3. Formatting association by depth visible at the vote
(win-odds change per SD, from the pooled formatting-only interaction
model; single-turn = \(e^{\gamma}-1\), multi-turn =
\(e^{\gamma+\delta}-1\)).}

{\def\LTcaptype{none} % do not increment counter
\begin{longtable}[]{@{}lcccl@{}}
\toprule\noalign{}
Feature & Single-turn & Multi-turn & Interaction \(\delta\) & Sig? \\
\midrule\noalign{}
\endhead
\bottomrule\noalign{}
\endlastfoot
\textbf{Bold} & \textbf{+30.1\%} & \textbf{+7.4\%} & −0.191 & Yes \\
Headers & +11.1\% & +6.8\% & −0.040 & Yes \\
Lists & +9.0\% & +7.8\% & −0.012 & No \\
Code blocks & +5.0\% & +0.8\% & −0.041 & Yes \\
Emoji & +2.4\% & +6.0\% & +0.035 & Yes \\
\end{longtable}
}

Bold changes the most, from +30.1\% win odds in single-turn battles to
+7.4\% in multi-turn battles, a reduction of about 75\% in the formatting-only
interaction model. This is a
difference between observed groups, not evidence that longer
conversations cause users to discount formatting. Users may continue for
reasons related to the prompt, their goals, or the quality of earlier
answers.

\begin{figure}[p]
\centering
\pandocbounded{\includegraphics[keepaspectratio,alt={Figure 2. Depth visible at the retained vote. Left: win-odds change per SD for each feature in single-turn (circle) vs multi-turn (diamond) battles. Right: the interaction with 95\% bootstrap CIs; grey is not significant after BH.}]{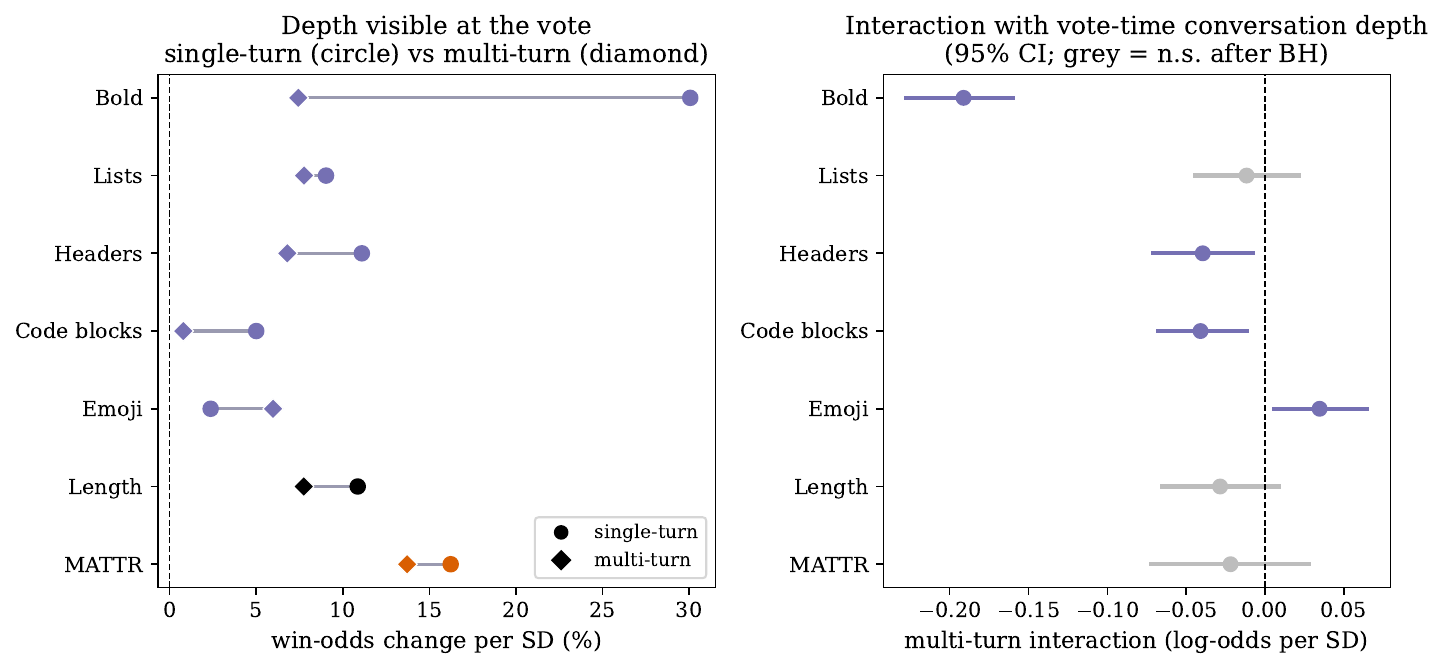}}
\caption{Depth visible at the retained vote. Left: win-odds
change per SD for each feature in single-turn (circle) vs multi-turn
(diamond) battles. Right: the interaction with 95\% bootstrap CIs; grey
is not significant after BH.}
\label{fig:vote-depth}
\end{figure}

Repeating the interaction analysis in the joint model gives a similar
contrast. Bold falls from +15.8\% to +3.1\% and headers from +8.6\% to
+3.5\%; both interactions remain significant after correction. Length
changes from +10.9\% to +7.7\%, and MATTR from +16.2\% to +13.7\%;
neither interaction is significant after correction. Separate
descriptive stratum fits are similar but are not used to derive these
tests. Section 4.4 shows that allowing observed topic-specific slopes
leaves the depth pattern similar.

\subsection{Subject Matter and Task
Type}\label{does-subject-matter-explain-the-formatting-association}

\paragraph{Subject matter.}

Allowing observed topic-specific slopes does not materially change the
bold association. Technical questions, for example, may invite both code
and heavier formatting. Since both compared models answer the same
prompt, topic alone cannot predict which side wins; the relevant question
is whether the formatting association
changes by topic. We therefore estimate the model separately within each
subject that has at least 2,500 battles. Within a stratum we then drop
models with fewer than 50 battles there, so several fitted samples are
smaller than the 2,500 selection threshold. The bold point estimate is positive
in \textbf{every} included subject, and most unadjusted 95\% bootstrap
intervals exclude zero (Table 4).

\begin{table}[htbp]
\textbf{Table 4. Bold association (odds change per SD) within each topic,
using capped contrasts (§3.2). Under the same specification, the
all-topic estimate is +22.3\%; Table 1 uses uncapped contrasts.}

\medskip
\centering
\begin{tabular}{@{}lr@{}}
\toprule\noalign{}
Topic & Bold \\
\midrule\noalign{}
Politics \& Government & +63.7\% \\
Law \& Justice & +43.9\% \\
Health \& Wellness \& Medicine & +36.0\% \\
Personal Development \& Career & +35.3\% \\
Arts & +34.8\% \\
Food \& Drink \& Cooking & +24.1\% \\
Daily Life \& Home \& Lifestyle & +22.7\% \\
Environment & +22.3\% \\
Society \& Social Issues & +21.6\% \\
Natural Science \& Technology & +20.9\% \\
Entertainment \& Travel \& Hobby & +19.1\% \\
Culture \& Cultural Geography & +18.0\% \\
Business \& Economics \& Finance & +16.9\% \\
Education & +15.0\% \\
\bottomrule\noalign{}
\end{tabular}
\end{table}

The size of the estimate varies, but smaller subject groups have wider
intervals, so the ordering should not be over-interpreted.

\begin{figure}[htbp]
\centering
\pandocbounded{\includegraphics[keepaspectratio,alt={Figure 3. Topic controls. Left: the bold association (win-odds change per SD) estimated within each topic, with unadjusted 95\% bootstrap CIs; the point estimate is positive in every subject. Right: the vote-time multi-turn interactions of §4.3 (circle) versus the same model with topic × formatting interactions added (diamond); the two nearly coincide.}]{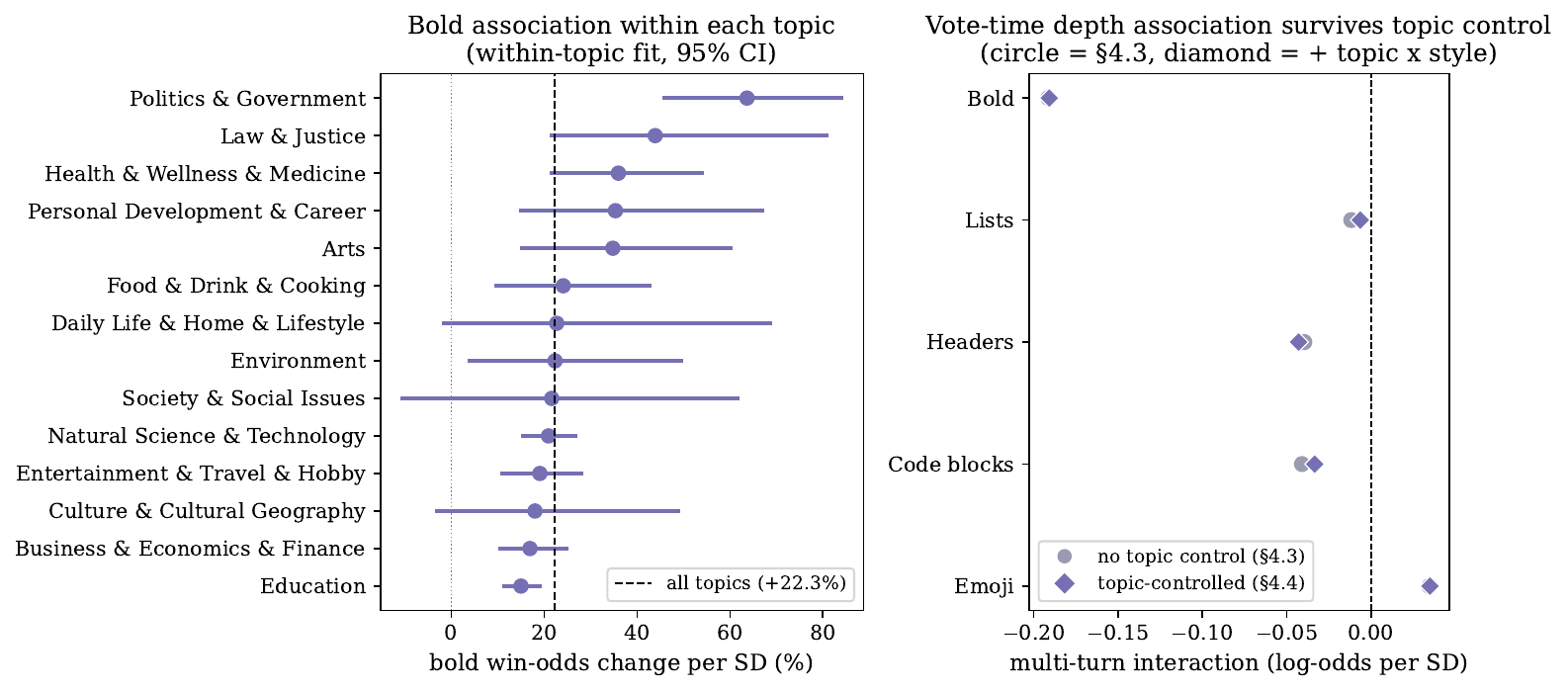}}
\caption{Topic controls. Left: the bold association (win-odds
change per SD) estimated within each topic, with unadjusted 95\%
bootstrap CIs. The point estimate is positive in every subject; the
intervals for three smaller subjects include zero. The dashed line is
the all-topic estimate under the same capped-contrast specification. Right: the vote-time multi-turn interactions
of §4.3 (circle) versus the same model with topic × formatting
interactions added (diamond); the two nearly coincide.}
\label{fig:topic-controls}
\end{figure}

\textbf{Topic adjustment leaves the multi-turn pattern similar.} After
allowing formatting associations to vary by topic, the multi-turn
interactions are almost unchanged: bold −0.191, headers −0.043, code
blocks −0.033, and emoji +0.035 are significant; lists −0.006 is not.
This does not rule out topic-related or unmeasured confounding, and the
analysis remains observational.

Topic here is subject matter, not task type (summarise, translate, write
code, give advice), which we examine next.

\paragraph{Task type.}\label{does-task-type-explain-the-formatting-association}

Task type is more directly related to formatting than subject matter. A
coding request invites code blocks, while a translation request may not.
We assign each opening prompt to one of ten broad task types using
ordered French keyword rules. The classifier has not been validated on
this release and about one-third of prompts fall into ``other.'' This is
therefore an exploratory proxy, not a precise task control.

\textbf{Bold remains positive in most task groups.} It is positive in
eight of nine reported groups. The unadjusted 95\% intervals exclude
zero for explanation (+26.7\%), writing (+15.1\%), code (+28.1\%), ideas
(+19.1\%), summarisation (+34.7\%), and advice (+38.9\%). The intervals
include zero for list/table, math, and translation (−11.8\%). These many
exploratory stratum-feature intervals are not multiplicity-adjusted.

The heterogeneous ``other'' category is not presented as a task stratum
because it has no coherent task interpretation; its size is one reason
we treat this analysis as exploratory rather than a complete task
control. Full intervals for every cell are retained in
\texttt{results/task\_results.json}.

\begin{table}[htbp]
\textbf{Table 5. Formatting associations (odds change per SD) within
each task. An asterisk marks an estimate whose unadjusted 95\% bootstrap
interval excludes zero.}

\medskip
\centering
\footnotesize
\setlength{\tabcolsep}{4pt}
\begin{tabular}{@{}lrrrrrr@{}}
\toprule\noalign{}
Task & Battles & Bold & Headers & Lists & Code blocks & Emoji \\
\midrule\noalign{}
explanation & 44,870 & +26.7\%* & +7.3\%* & +14.4\%* & −1.0\% &
+0.9\% \\
writing & 12,587 & +15.1\%* & +9.6\%* & +3.8\% & −6.1\%* &
+11.5\%* \\
code & 9,749 & +28.1\%* & +6.3\% & +2.5\% & +13.1\%* & +0.6\% \\
ideas & 4,408 & +19.1\%* & +13.3\%* & +3.2\% & +6.4\% & −2.1\% \\
list/table & 2,956 & +7.8\% & +16.9\%* & −3.3\% & +16.8\%* &
+13.1\%* \\
summarisation & 2,943 & +34.7\%* & +10.2\% & +2.0\% & −4.5\% &
+5.7\% \\
advice & 2,618 & +38.9\%* & −0.7\% & −1.6\% & −1.5\% & −0.2\% \\
translation & 2,451 & −11.8\% & +3.0\% & +12.1\% & +23.2\%* &
+5.5\% \\
math & 1,472 & +16.3\% & +1.5\% & +4.7\% & +10.3\% & +7.2\% \\
\bottomrule\noalign{}
\end{tabular}
\end{table}

\textbf{Bold survives the task split; the other features do not behave
uniformly.} Bold is positive in eight of the nine reported groups and
its interval excludes zero in six of them; the exception is translation
(−11.8\%), where markdown emphasis has no obvious job to do. Headers and
lists are positive on average but reach conventional significance in
only a few groups each, which is what one expects of features whose
signal is largely shared. Code blocks are the clearest case of
heterogeneity hidden by a pooled estimate: positive for code (+13.1\%),
translation (+23.2\%), and list/table (+16.8\%), but near zero or
negative in prose tasks such as writing (−6.1\%) and summarisation
(−4.5\%). Averaging across mostly non-coding battles is therefore a
plausible explanation for the null pooled estimate in §4.1. A validated
task taxonomy is still needed before any of this is more than
suggestive.

\subsection{Robustness Checks}\label{robustness-to-arena-mode-and-temporal-dependence}

\paragraph{Arena mode and temporal dependence.}

The main results survive two checks for how battles were sampled. First,
72\% of decisive battles use random model pairs, while 19\% use pairs
selected by the user. The random-pair analysis contains 98,306 battles
for the formatting model and 91,070 for the full model. Bold is +20.2\%
in the formatting model and +12.4\% in the full model; MATTR is +16.0\%.
These are close to the full-sample estimates of +19.0\%, +11.0\%, and
+16.8\%. Headers and length move more, to +5.1\% and +5.0\%, reinforcing
the conclusion that their individual estimates are less stable.

Second, we resample the data by calendar week to account for changes in
the model roster and user population over time. The 95\% intervals
remain above zero: formatting-only bold +19.0\% {[}+15.2, +23.6{]},
full-model bold +11.0\% {[}+8.7, +13.3{]}, and MATTR +16.8\% {[}+14.4,
+19.1{]}. The release has no user identifier, so dependence among
battles from the same user cannot be tested.

\paragraph{Lexical diversity.}\label{testing-the-lexical-diversity-result}

The MATTR association persists under sensitivity checks for answer
length, the 50-token window, function words, and proper names. Responses
are long enough for this measure: the median is 691 output tokens (IQR
380--1,138), and only 4.9\% fall below the window and are excluded.
MATTR is nearly unrelated to length overall (Spearman +0.05) and within
length quartiles (−0.04 to +0.11).

The association is stronger for above-median-length answers (+16.5\%)
than for shorter answers (+4.5\%), while length shows the opposite
pattern (+6.3\% for longer answers and +44.6\% for shorter ones). Among
short-answer battles, a larger length contrast is more strongly
associated with winning. Among already long answers, additional length
matters less and lexical diversity carries more of the measured signal.
MATTR remains positive in both groups.

We also checked whether the result depends on the chosen diversity
measure or simply captures names and technical terms. In otherwise
identical joint models on each metric's available-case support, the
estimate remains positive: \textbf{+16.8\% for MATTR, +12.7\% for MTLD
(McCarthy \& Jarvis, 2010), +11.7\% after removing French function
words, and +13.3\% after excluding capitalised tokens as a rough proxy
for proper names}. MATTR
and MTLD are closely related in these data (Spearman 0.94). We therefore
describe lexical diversity as a stable correlate, not proven vocabulary
``richness,'' because it may still capture topical specificity.

\textbf{Table 6. Alternative lexical-diversity specifications (odds
change per SD; metric-specific available-case support).}

{\def\LTcaptype{none}
\begin{longtable}[]{@{}lrr@{}}
\toprule\noalign{}
Specification & Battles & Odds change \\
\midrule\noalign{}
\endhead
\bottomrule\noalign{}
\endlastfoot
MATTR & 127,092 & +16.8\% \\
MTLD & 127,092 & +12.7\% \\
MATTR without French function words & 119,503 & +11.7\% \\
MATTR excluding capitalised tokens & 125,364 & +13.3\% \\
\end{longtable}
}

\subsection{Production Rankings and External
Benchmarks}\label{production-rankings-and-external-benchmarks}

The live production leaderboard shows how far style control can move
prominent models. Compar:IA enables style control by default
and describes it as removing the influence of response length and
formatting (Compar:IA, 2026a). In the live snapshot observed on 27 July
2026, switching that control changes several prominent positions
substantially.

\begin{table}[htbp]
\textbf{Table 7A. Illustrative ranks on the live Compar:IA leaderboard,
observed 27 July 2026. These are production ranks, not ranks
reconstructed from the research release.}

\medskip
\centering
\begin{tabular}{@{}lrrl@{}}
\toprule\noalign{}
Model & Raw & Style-controlled & Epoch Capabilities \\
 & live rank & live rank & Index coverage \\
\midrule\noalign{}
GPT-5.3 & 47 & 1 & Not matched \\
Mistral Medium 2508 & 2 & 28 & Not matched \\
Gemini 3.1 Flash Lite & 4 & 4 & Not matched \\
Gemini 2.5 Flash & 5 & 12 & 140.33 \\
Gemini 3.1 Pro & 15 & 27 & 154.90 \\
\bottomrule\noalign{}
\end{tabular}
\end{table}

The shift makes the controlled ranking look more plausible if one begins
with the expectation that GPT-5.3 should lead and Mistral Medium should
not. That is useful \textbf{face-validity evidence}, not an independent
validation: the expectation itself comes from prior beliefs and other
evaluations, and style control does not resolve every surprising
ordering. Gemini 3.1 Flash Lite remains above Gemini 3.1 Pro, while
Gemini 2.5 Flash also remains above it. The live page displayed a
counter of roughly 242,000 reactions and 112 ranked models when
observed. That counter includes ties, whereas this paper retains only
decisive French battles; the paper also uses an earlier pinned research
release. Its retained frame is therefore smaller, at 137,293 battles,
and its analyses use 116 models meeting the minimum-battle threshold.
The production and research ranks must not be compared as if they came
from the same snapshot and filtering rule.

The main external question is whether presentation control changes
agreement with benchmarks that do not use arena preferences. We compare
the raw, formatting-controlled, and full joint-controlled Compar:IA
rankings on the same 127,092 battles and 116 models used in §4.2. For
every external benchmark, all three correlations use exactly the same
matched model versions. We match exact identifiers or a small set of
manually audited aliases for the same model build; nearby releases,
model families, reasoning levels, and tool configurations are not
merged. We require at least ten matches, calculate Spearman rank
correlations, and bootstrap the change from the raw ranking 10,000
times.

We use audited statistics derived from the Epoch AI archive retrieved on
27 July 2026 (Epoch AI, 2026). The source URL is mutable and now returns
different bytes; the repository preserves the original SHA-256, archive
manifest, matched scores, and exclusions, but not the original archive
payload. The sensitivity analysis is therefore verifiable from retained
derived scores and audits, but it cannot be independently re-downloaded
end to end from the original payload. The Epoch Capabilities Index is
the primary broad comparison. GPQA
Diamond, FrontierMath, LiveBench, ARC-AGI-2, SciCode, Aider Polyglot,
and SWE-bench Verified provide domain-specific checks (Rein et al.,
2023; Glazer et al., 2024; White et al., 2025; Chollet et al., 2025;
Tian et al., 2024; Aider, 2025; Jimenez et al., 2024). These benchmarks partly measure
different capabilities, and models show some specialisation across them
(Ho et al., 2025), so disagreement among the comparisons is expected.

\begin{table}[htbp]
\textbf{Table 7B. Spearman correlation with non-arena capability
benchmarks. Bold marks the highest point correlation in each row, not a
statistically superior ranking.}

\medskip
\centering
\small
\setlength{\tabcolsep}{4pt}
\begin{tabular}{@{}lrrrrl@{}}
\toprule\noalign{}
Capability benchmark & Matches & Raw & Formatting- & Full joint- & Formatting-controlled \\
 & & & controlled & controlled & minus raw (95\% CI) \\
\midrule\noalign{}
Epoch Capabilities Index & 38 & \textbf{0.717} & 0.703 & 0.635 & −0.014
{[}−0.079, +0.054{]} \\
GPQA Diamond & 32 & \textbf{0.753} & 0.741 & 0.664 & −0.011 {[}−0.063,
+0.035{]} \\
FrontierMath & 13 & \textbf{0.699} & 0.655 & 0.534 & −0.044 {[}−0.223,
0.000{]} \\
LiveBench & 17 & \textbf{0.419} & 0.277 & 0.358 & −0.142 {[}−0.485,
+0.097{]} \\
ARC-AGI-2 & 10 & \textbf{0.537} & 0.488 & 0.303 & −0.049 {[}−0.367,
+0.209{]} \\
SciCode & 12 & \textbf{0.413} & 0.399 & 0.336 & −0.014 {[}−0.149,
+0.073{]} \\
Aider Polyglot & 10 & 0.382 & 0.345 & \textbf{0.539} & −0.036 {[}−0.314,
+0.199{]} \\
\bottomrule\noalign{}
\end{tabular}
\end{table}

Formatting control has a lower point correlation than the raw ranking on
all seven eligible capability benchmarks. Every formatting-controlled
minus raw interval includes zero, so the data do not establish a
difference in either direction. FrontierMath is the borderline case: its
interval ends exactly at zero ({[}−0.223, 0.000{]}). This aggregate test
does not cover the production examples evenly: GPT-5.3, Mistral Medium
2508, and Gemini 3.1 Flash Lite are absent from the Epoch Capabilities
Index match, while Gemini 2.5 Flash and Gemini 3.1 Pro are included. The
benchmark therefore cannot test whether the largest live shifts move
those omitted models toward capability. Full joint control is lower on
six benchmarks and higher only on Aider Polyglot, where the matched
sample is ten models and the difference is highly uncertain. Kendall
correlations and leave-one-provider-out checks do not change the overall
interpretation. SWE-bench Verified has only nine matching model versions
and is excluded by the stated minimum-overlap rule.

\begin{figure}[htbp]
\centering
\pandocbounded{\includegraphics[width=0.80\textwidth,keepaspectratio,alt={Figure 4. Changes in Spearman correlation with seven non-arena capability benchmarks. Formatting-controlled and full joint-controlled estimates are shown relative to the raw Compar:IA ranking with paired 95 percent bootstrap intervals.}]{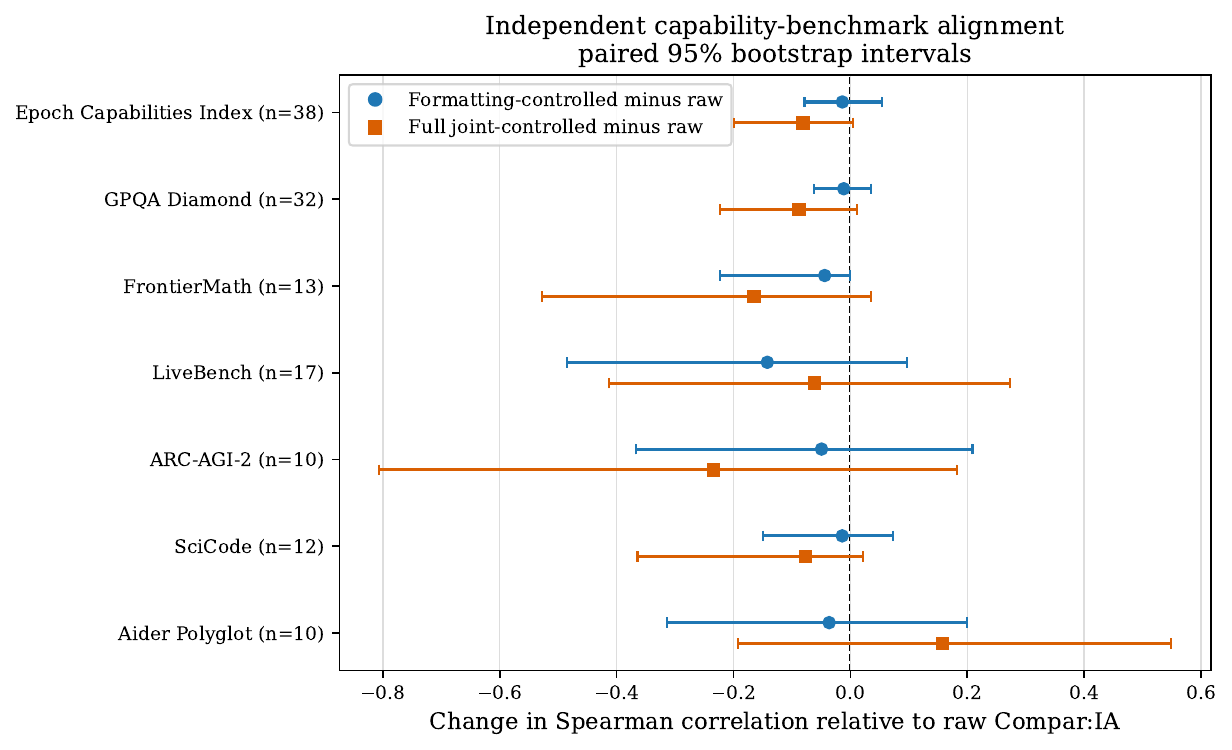}}
\caption{Change in Spearman correlation relative to raw
Compar:IA for non-arena capability benchmarks. Points show
formatting-controlled and full joint-controlled rankings; lines are
paired 95\% bootstrap intervals.}
\label{fig:external-alignment}
\end{figure}

LMArena is a secondary comparison because it is another human-preference
arena rather than an independent capability benchmark. We pin its 16
July 2026 Text Arena revision and match exact public identifiers (Arena
Team, 2026).

\begin{table}[htbp]
\textbf{Table 7C. Spearman correlation with LMArena preference rankings.
Bold marks the highest point correlation in each row.}

\medskip
\centering
\begin{tabular}{@{}lrrrr@{}}
\toprule\noalign{}
LMArena preference ranking & Matches & Raw & Formatting-controlled & Full joint-controlled \\
\midrule\noalign{}
Raw, overall & 49 & 0.792 & \textbf{0.800} & 0.710 \\
Style-controlled, overall & 49 & 0.768 & \textbf{0.808} & 0.735 \\
Raw, French & 40 & 0.779 & \textbf{0.796} & 0.698 \\
Style-controlled, French & 40 & 0.701 & \textbf{0.773} & 0.693 \\
\bottomrule\noalign{}
\end{tabular}
\end{table}

Formatting control has a slightly higher point correlation in all four
LMArena comparisons, from +0.008 against the raw overall ranking to
+0.072 against the style-controlled French ranking. All four
formatting-controlled intervals include zero. Full joint control is
lower in all four; its interval excludes zero only against the raw
overall LMArena ranking (−0.082 {[}−0.162, −0.010{]}). The capability
and preference comparisons therefore point in different directions, but
neither supplies broad, decisive evidence. They show that presentation
adjustment changes what the ranking tracks; they do not show which
ranking is better.

\section{Discussion}\label{discussion}

\subsection{What the Associations Show}\label{what-the-results-show}

Presentation is associated with arena votes, but not as a set of cleanly
separable effects. Longer answers also tend to use more bold text and
lists. The model must divide this shared association among correlated
features, so the estimate for any one of them depends on what else is
included. Among the same set of battles, for example, bold is associated
with +20.4\% win odds in the formatting-only model and +11.0\% after
length and language are added.

Two associations are less sensitive to these changes: \textbf{bold
usage} and \textbf{comparatively length-robust lexical diversity},
measured with the moving-average type-token ratio (MATTR). MATTR measures
vocabulary variety within rolling windows and is nearly unrelated to
total answer length in these data. We avoid calling it vocabulary
“richness,” however, because technical terms and topic-specific
language can also increase the measure. Readability contributes little
once length is included.

The associations also differ across observed conversation depth. In the
joint model, bold falls from +15.8\% in single-turn battles to +3.1\% in
multi-turn battles, while MATTR changes from +16.2\% to +13.7\%. The
bold interaction is statistically significant after correction; the
MATTR interaction is not. This contrast motivates the signalling
interpretation developed in §5.5.

Nor do the external comparisons identify a superior ranking. Some large
movements on the production leaderboard, most notably for GPT-5.3 and
Mistral Medium, may look plausible given prior expectations. But the
aggregate capability comparisons do not cover those models. Among the
models they do cover, formatting control has a lower point correlation
with every eligible capability benchmark and a slightly higher point
correlation with all four LMArena preference rankings. None of the
formatting-versus-raw differences is statistically decisive. Adjustment
clearly changes what the ranking tracks; these data do not show that it
tracks capability better.

\subsection{Why Association Is Not
Causation}\label{why-association-is-not-causation}

Two explanations fit the results. Presentation may sway users
independently of content, in which case style control removes a bias.
Alternatively, stronger models may produce clearer structure as part of
a better answer, in which case style control removes useful quality
signal. The data cannot determine how much each explanation contributes.
Three descriptive checks show why.

\textbf{Test 1: Preferred models format more.} A model's average use of
bold, lists, and headers correlates with its raw rating at Pearson r =
0.60 (Spearman 0.67). After style control, the correlation is r = 0.46.
This pattern is consistent with presentation carrying both quality and
presentation-specific preference, but it cannot separate them because
the rating comes from the same votes.

\textbf{Test 2: The association varies by rating tier.} We split models
into tiers using their raw ratings and re-estimate the formatting
associations within each tier:

\textbf{Table 8. Formatting association by model-pair tier (odds change
per SD).}

{\def\LTcaptype{none}
\begin{longtable}[]{@{}lccc@{}}
\toprule\noalign{}
Feature & Bottom & Middle & Top \\
\midrule\noalign{}
\endhead
\bottomrule\noalign{}
\endlastfoot
Bold & +20.3\% & +12.2\% & +13.5\% \\
Lists & +13.8\% & +8.8\% & −5.4\% \\
Headers & +10.5\% & +1.0\% & +13.7\% \\
N battles & 32,439 & 13,691 & 15,928 \\
\end{longtable}
}

Bold has its largest association in the bottom tier (+20.3\%, compared
with +13.5\% in the top tier). This could indicate a stronger
presentation premium among lower-rated models, but the tiers are
themselves defined by vote-based ratings, so the comparison is not
causal.

\textbf{Test 3: Models that format more move down more.} Rating change
after style control correlates at r = −0.98 with average formatting.
This confirms that the adjustment is operating as designed, but it does
not tell us whether the removed signal was bias or quality.

Together, these checks point to the same unresolved distinction.
Presentation may be part of answer quality, a separate influence on
preference, or both. Accounting for model identity removes stable
model-level differences from the estimated feature associations, but it
does not isolate within-model causal effects. Neither the raw nor the
adjusted ranking is a definitive measure of quality. Distinguishing the
two explanations requires an experiment that varies presentation while
holding content fixed.

\textbf{Winner-flipping battles provide another descriptive check.} A
“winner-flipping” battle is one in which the raw and
formatting-controlled ratings imply different winners for the model
pair. This occurs in \textbf{7,273 of 137,113 battles (5.3\%)}. Among
these battles, the actual vote winner uses more bold, lists, and headers
in \textbf{51.6\%} of cases, compared with \textbf{40.9\%} for the
loser. The models most often involved include heavy formatters that move
down after control (mistral-large-2512, mistral-medium-2508) and models
that frequently face them (llama-3.1-405b, claude-4-6-sonnet).

This release carries no per-message reaction data, so user-reported
reasons for a vote, such as a ``clear formatting'' attribute, cannot be
linked to these battles.

\subsection{Limitations}\label{limitations}

\textbf{Causal identification.} This is an observational study.
Formatting, length, and diversity vary with prompt difficulty, task
type, correctness, refusals, and conversation history. Model controls do
not remove these unmeasured differences, and correlated features such as
length, bold, and lists cannot be cleanly separated. The coefficients
should therefore not be read as the effects of adding one feature to an
otherwise unchanged answer. Conversation depth has the same problem:
users choose whether to continue based on earlier answers, the task, and
their goals. A randomised presentation experiment is needed for causal
claims.

\textbf{Measurement.} MATTR captures vocabulary variety, not language
quality as a whole. The stress tests in §4.5 rule out several simple
explanations for its association: total length, the 50-token window,
function words, and proper names. It may still reflect topic-specific
language, and we did not test HD-D. Two of the three readability
formulas, Coleman-Liau and Flesch-Kincaid, were calibrated for English
rather than French; only REL is French-specific. The task classifier is
an unvalidated keyword proxy that will miss implicit tasks. Fluency is
also absent: CamemBERT pseudo-perplexity required a GPU and was not
recomputed for this release. It added nothing beyond length and
readability in an earlier export, but that result may not carry over.

The formatting model is additive. Section 4.1 reports the ten pairwise
interactions and finds none that survives correction, but sparse and
strongly correlated features mean that check has limited power against
particular bundles.

\textbf{Dependence and generalisability.} The release has no user or
session identifier. Although we retain one vote per conversation and
the headline results survive random-pair restriction and weekly block
resampling, we cannot account for one person contributing several
conversations or estimate the resulting change in uncertainty. The data
also come from one platform with a self-selected and demographically
unobserved population. Cross-platform ranking comparisons are not a
replication of the battle-level analysis; that requires response-level
data from another arena or population.

\textbf{External validation.} Benchmark overlap is limited to 10--38
model versions for capability benchmarks and 40--49 for LMArena. Even
exact public model names can hide differences in system prompts,
inference budgets, scaffolds, provider settings, and evaluation dates;
related model variants are not independent. The largest live production
shifts are especially poorly covered: GPT-5.3, Mistral Medium 2508, and
Gemini 3.1 Flash Lite are absent from the Epoch Capabilities Index match,
and the live examples come from a different Compar:IA snapshot than the
paper analysis. We preserve the source hash, audit same-build matches,
report exclusions, and run provider-omission checks where possible, but
these comparisons remain uncertain and cannot establish that the
largest live movements improve alignment with capability.

\subsection{What This Means for
Arenas}\label{implications-for-arena-design}

\begin{enumerate}
\def\labelenumi{\arabic{enumi}.}
\item
  \textbf{Publish raw and adjusted rankings together.} The distance
  between them shows how sensitive a model's position is to measured
  presentation; it does not reveal which ranking is more correct.
  Heavy-formatting models such as mistral-large-2512 and gpt-oss-120b
  fall after adjustment, but that alone does not prove that their raw
  ranks were inflated.
\item
  \textbf{State what the adjustment removes.} A control for length and
  markdown may also remove part of what those features share with answer
  completeness. That can be a defensible design choice, but it should be
  described as a choice rather than as the recovery of a uniquely
  correct ranking.
\item
  \textbf{Run controlled presentation experiments.} Arenas could
  randomly vary or normalise the formatting of otherwise identical
  content. Such an experiment would isolate whether markup itself
  changes votes and whether the effect differs by conversation context.
\item
  \textbf{Watch the incentives rankings create.} If arena positions
  influence model development, an association between visible style and
  winning may reward more conspicuous presentation. The present study
  identifies that possibility but cannot measure its trade-off with
  substantive quality.
\end{enumerate}

\subsection{Signalling as a Hypothesis Across Conversation
Depth}\label{signaling-perspective}

Signalling, often called cueing in educational psychology, offers one
way to interpret the single-turn versus multi-turn result. Signals such as headings, typographical emphasis, and visual
pointers highlight important material or make its organisation easier
to see without necessarily adding new content. In a meta-analysis of 103
studies and 12,201 participants, signalling improved retention and
transfer on average, reduced reported cognitive load, and directed more
visual attention toward learning-relevant material (Schneider et al.,
2018). That literature concerns learning from instructional media rather than
choosing between competing answers, so it transfers to an arena only
with qualification.

Markdown nevertheless resembles an organisational cue in one specific
respect: bold text, headings, and code blocks can make an answer's
structure visible. If such cues help voters locate an answer's
priorities, their additional value may be greatest when little other
evidence about the models is available. As a conversation continues,
voters also observe how each model follows instructions, responds to
corrections, and maintains context. The smaller multi-turn associations
for bold and headers are compatible with cues carrying less additional
weight in that richer setting. We should be careful about the step being
taken here, because the signalling literature does not supply this
prediction. The same meta-analysis tested the nearest available
moderator, learner prior knowledge, and did not find that it moderated
the signalling effect (Schneider et al., 2018). How much a learner
already knows and how much of a conversation a voter has already read
are different variables, so this is not a contradiction, but the
extension from one to the other is ours. The relative stability of MATTR is also
compatible with a feature whose association emerges through the
language of the answer rather than its visual prominence.

One way to frame the contrast is as a hypothesis about possible
\textbf{levels of reading}: visually apparent cues and properties that
emerge through language may enter a judgement differently. This is a
hypothesis, and turn count does not measure attention: single-turn votes
are not necessarily shallow readings. The data raise this as a question
for experiment; they do not establish the mechanism.

The interpretation does produce a direct prediction. Holding answer
content fixed, randomly adding or removing organisational cues should
change choices more in single-turn comparisons than in multi-turn
comparisons if cueing explains the observed heterogeneity. Such an
experiment could combine votes with consented measures of inspection,
such as dwell time or scrolling, to test whether the cues actually guide
attention. Until then, signalling remains a hypothesis that connects
the depth result to a broader literature.

\section{Conclusion}\label{conclusion}

Presentation is associated with Compar:IA votes, but the data do not
support interpreting bold, lists, headers, and length as separate
preference effects. These features often occur together,
and their estimates change as the model separates their shared signal.
Bold usage and comparatively length-robust vocabulary variety (MATTR)
are the two most stable correlates. Readability adds little. The
pooled-versus-model-controlled comparison also shows that stable
differences between models account for much of the apparent language
association.

The associations also differ by observed conversation depth. Bold,
headers, and code blocks have smaller associations in multi-turn
conversations, whereas length and MATTR are more stable. Because users
choose whether to continue, this difference is descriptive, but it
motivates the testable cueing hypothesis developed in the discussion.

Adjustment can move prominent models by dozens of ranks, but movement is
not validation. Formatting control correlates slightly less with seven
non-arena capability benchmarks and slightly more with four LMArena
preference rankings; none of the raw-versus-formatting differences is
decisive. The data show that presentation is associated with votes and
changes what a controlled arena ranking captures. They do not show that
the adjusted ranking is generally more
accurate, or whether presentation represents bias, answer quality, or
both.

Given these limits, publishing the raw ranking beside prespecified
adjusted rankings would make the sensitivity transparent. Two next
steps would make the evidence substantially
stronger: replace the rough task proxy with a validated classifier, and
randomly vary presentation while holding answer content fixed.

\section*{Data and Code Availability}\label{data-and-code-availability}

The source dataset is available through the gated Hugging Face
repository \nolinkurl{ministere-culture/comparia-fr-arena} at the
immutable revision reported in §2.2. The paper repository distributes the
text-free derived tables, result files, figures, LaTeX source, a locked
Python environment, tests, and a SHA-256 artifact manifest. Raw prompts
and conversation text are not redistributed. The external Epoch source
URL is mutable; the repository includes the audited source hash, archive
manifest, model-match audit, and derived scores, but not the original
archive payload.

Repository: \url{https://github.com/simonaszilinskas/style-control-analysis}.
The exact state used for this paper is archived on Zenodo as release v1.0:
\url{https://doi.org/10.5281/zenodo.23043602}.

\section*{Ethics and Privacy}\label{ethics-and-privacy}

This is a secondary analysis of an existing research release; no new
user interaction or data collection was conducted. The gated source
contains user-generated text. The distributed analysis tables retain
derived measurements, model identifiers, outcomes, timestamps, topics,
task proxies, and opaque comparison identifiers, but no prompt or
response text. The source release's access conditions and licences
continue to apply.

\section*{Acknowledgements}\label{acknowledgements}

We gratefully acknowledge the students who participated in the hackathon
held at Université Paris Dauphine-PSL in early June 2026, led by
coauthor Christophe Benavent. Their analyses of response style in
the Compar:IA dataset informed several of the questions examined in this
paper.

\section*{References}\label{references}

\begingroup
\raggedright

\begin{itemize}
\tightlist
\item
  Aider. (2025). \emph{Aider LLM Leaderboards}.
  \url{https://aider.chat/docs/leaderboards/} Last updated 20 November
  2025; accessed 29 September 2026. Polyglot scores used here come from
  the Epoch AI archive.
\item
  Arena Team. (2026). \emph{Arena Leaderboard Dataset}.
  \url{https://arena.ai/blog/arena-leaderboard-dataset/} Dataset:
  \url{https://huggingface.co/datasets/lmarena-ai/leaderboard-dataset},
  revision \texttt{afed939e10281b660a4369206ca505b2bf5e0208}, leaderboard
  date 16 July 2026.
\item
  Benjamini, Y., \& Hochberg, Y. (1995). Controlling the false discovery
  rate: A practical and powerful approach to multiple testing.
  \emph{Journal of the Royal Statistical Society: Series B}, 57(1),
  289--300. \url{https://doi.org/10.1111/j.2517-6161.1995.tb02031.x}
\item
  Bradley, R. A., \& Terry, M. E. (1952). Rank analysis of incomplete
  block designs: I. The method of paired comparisons. \emph{Biometrika},
  39(3/4), 324--345. \url{https://doi.org/10.2307/2334029}
\item
  Chollet, F., Knoop, M., Kamradt, G., Landers, B., \& Pinkard, H.
  (2025). ARC-AGI-2: A new challenge for frontier AI reasoning systems.
  \emph{arXiv:2505.11831}. \url{https://arxiv.org/abs/2505.11831}
\item
  Coleman, M., \& Liau, T. L. (1975). A computer readability formula
  designed for machine scoring. \emph{Journal of Applied Psychology},
  60(2), 283--284. \url{https://doi.org/10.1037/h0076540}
\item
  Compar:IA. (2026a). \emph{Classement - Compar:IA, le comparateur d'IA
  conversationnelles} (Méthodologie tab).
  \url{https://arene.comparia.beta.gouv.fr/ranking} Observed 27 July 2026
  at \url{https://comparia.beta.gouv.fr/ranking}, which has since moved;
  accessed at the current address 29 September 2026.
\item
  Compar:IA \& Ministère de la Culture. (2026b).
  \emph{comparia-fr-arena: French-language conversations and human
  preferences} (gated).
  \url{https://huggingface.co/datasets/ministere-culture/comparia-fr-arena}
  Revision \texttt{8cd6488c5d0c3b8dfcb9339d11ae9624c84359be}, accessed 8
  July 2026. Open Licence 2.0 (Etalab) and CC-BY-4.0.
\item
  Covington, M. A., \& McFall, J. D. (2010). Cutting the Gordian knot:
  The moving-average type--token ratio (MATTR). \emph{Journal of
  Quantitative Linguistics}, 17(2), 94--100.
  \url{https://doi.org/10.1080/09296171003643098}
\item
  Dubois, Y., Galambosi, B., Liang, P., \& Hashimoto, T. B. (2024).
  Length-controlled AlpacaEval: A simple way to debias automatic
  evaluators. \emph{arXiv:2404.04475}.
  \url{https://arxiv.org/abs/2404.04475}
\item
  Efron, B. (1979). Bootstrap methods: Another look at the jackknife.
  \emph{Annals of Statistics}, 7(1), 1--26.
  \url{https://doi.org/10.1214/aos/1176344552}
\item
  Epoch AI. (2026). \emph{AI Benchmarking Hub}.
  \url{https://epoch.ai/benchmarks/use-this-data} Snapshot retrieved 27 July
  2026; archive SHA-256:\\
  \nolinkurl{08ed76781fe84ce0cf6c80500cdae7ed347aaf71b7ac74cd016d31198424f3e4}.
\item
  Glazer, E., Erdil, E., Besiroglu, T., Chicharro, D., Chen, E.,
  Gunning, A., Olsson, C. F., et al.~(2024). FrontierMath: A benchmark
  for evaluating advanced mathematical reasoning in AI.
  \emph{arXiv:2411.04872}. \url{https://arxiv.org/abs/2411.04872}
\item
  Ho, A., Denain, J.-S., Atanasov, D., Albanie, S., \& Shah, R. (2025).
  A Rosetta Stone for AI benchmarks. \emph{arXiv:2512.00193}.
  \url{https://arxiv.org/abs/2512.00193}
\item
  Jimenez, C. E., Yang, J., Wettig, A., Yao, S., Pei, K., Press, O., \&
  Narasimhan, K. (2024). SWE-bench: Can language models resolve
  real-world GitHub issues? \emph{Proceedings of ICLR 2024}.
  \url{https://arxiv.org/abs/2310.06770}
\item
  Kandel, L., \& Moles, A. (1958). Application de l'indice de Flesch à
  la langue française. \emph{Cahiers d'Études de Radio-Télévision}, 19,
  253--274.
\item
  Kincaid, J. P., Fishburne, R. P., Rogers, R. L., \& Chissom, B. S.
  (1975). \emph{Derivation of new readability formulas (Automated
  Readability Index, Fog Count and Flesch Reading Ease Formula) for Navy
  enlisted personnel}. Research Branch Report 8-75, Chief of Naval
  Technical Training, Naval Air Station Memphis.
\item
  Li, T., Angelopoulos, A. N., \& Chiang, W.-L. (2024). Does style
  matter? Disentangling style and substance in Chatbot Arena.
  \emph{LMSYS Org blog}.
  \url{https://www.lmsys.org/blog/2024-08-28-style-control/}
\item
  McCarthy, P. M., \& Jarvis, S. (2010). MTLD, vocd-D, and HD-D: A
  validation study of sophisticated approaches to lexical diversity
  assessment. \emph{Behavior Research Methods}, 42(2), 381--392.
  \url{https://doi.org/10.3758/BRM.42.2.381}
\item
  Rein, D., Hou, B. L., Stickland, A. C., Petty, J., Pang, R. Y.,
  Dirani, J., Michael, J., \& Bowman, S. R. (2023). GPQA: A
  graduate-level Google-proof Q\&A benchmark. \emph{arXiv:2311.12022}.
  \url{https://arxiv.org/abs/2311.12022}
\item
  Saito, K., Wachi, A., Wataoka, K., \& Akimoto, Y. (2023). Verbosity
  bias in preference labeling by large language models.
  \emph{arXiv:2310.10076}. \url{https://arxiv.org/abs/2310.10076}
\item
  Schneider, S., Beege, M., Nebel, S., \& Rey, G. D. (2018). A
  meta-analysis of how signaling affects learning with media.
  \emph{Educational Research Review}, 23, 1--24.
  \url{https://doi.org/10.1016/j.edurev.2017.11.001}
\item
  Singhal, P., Goyal, T., Xu, J., \& Durrett, G. (2024). A long way to
  go: Investigating length correlations in RLHF. \emph{Proceedings of
  COLM 2024}. \url{https://openreview.net/forum?id=G8LaO1P0xv}
\item
  Tian, M., Gao, L., Zhang, S. D., Chen, X., Fan, C., Guo, X., Haas, R.,
  et al.~(2024). SciCode: A research coding benchmark curated by
  scientists. \emph{NeurIPS 2024 Datasets and Benchmarks}.
  \url{https://arxiv.org/abs/2407.13168}
\item
  White, C., Dooley, S., Roberts, M., Pal, A., Feuer, B., Jain, S.,
  Shwartz-Ziv, R., et al.~(2025). LiveBench: A challenging,
  contamination-limited LLM benchmark. \emph{Proceedings of ICLR 2025}.
  \url{https://proceedings.iclr.cc/paper_files/paper/2025/hash/e4a46394ba5378b3f9a186a5b4c650d1-Abstract-Conference.html}
\item
  Wu, M., \& Aji, A. F. (2025). Style over substance: Evaluation biases
  for large language models. \emph{Proceedings of COLING 2025},
  297--312. \url{https://aclanthology.org/2025.coling-main.21/}
\item
  Zheng, L., Chiang, W.-L., Sheng, Y., Zhuang, S., Wu, Z., Zhuang, Y.,
  Lin, Z., Li, Z., Li, D., Xing, E. P., Zhang, H., Gonzalez, J. E., \&
  Stoica, I. (2023). Judging LLM-as-a-judge with MT-Bench and Chatbot
  Arena. \emph{NeurIPS 2023 Datasets and Benchmarks}.
  \url{https://arxiv.org/abs/2306.05685}
\end{itemize}

\endgroup

\clearpage
\appendix
\section{Analysis Pipeline}\label{appendix-a-analysis-pipeline}

The core results are reproducible from the battle table
\nolinkurl{data/fr_battles.parquet}, built from
\nolinkurl{comparia-fr-arena} by
\nolinkurl{src/build_fr_arena.py}. \nolinkurl{run.py} exposes
\texttt{core}, \texttt{extended}, and \texttt{full} profiles with
explicit raw-data prerequisites; the repository README documents the
exact commands (scripts live in \texttt{src/}, outputs in
\texttt{results/}).

{\def\LTcaptype{none} % do not increment counter
\footnotesize
\setlength{\tabcolsep}{3pt}
\renewcommand{\arraystretch}{0.95}
\begin{longtable}[]{@{}
  >{\raggedright\arraybackslash}p{(\linewidth - 4\tabcolsep) * \real{0.4600}}
  >{\raggedright\arraybackslash}p{(\linewidth - 4\tabcolsep) * \real{0.1300}}
  >{\raggedright\arraybackslash}p{(\linewidth - 4\tabcolsep) * \real{0.4100}}@{}}
\toprule\noalign{}
\begin{minipage}[b]{\linewidth}\raggedright
Script
\end{minipage} & \begin{minipage}[b]{\linewidth}\raggedright
Section
\end{minipage} & \begin{minipage}[b]{\linewidth}\raggedright
Output
\end{minipage} \\
\midrule\noalign{}
\endhead
\bottomrule\noalign{}
\endlastfoot
\texttt{analyze\_core.py} & §4.1 & formatting Bradley-Terry model,
rank changes, position bias \\
\texttt{formatting\_interactions.py} & §4.1 & pairwise formatting
interactions and likelihood-ratio test \\
\texttt{linguistic\_analysis.py} & §4.2 & joint
formatting+length+linguistic model \\
\texttt{leaderboard\_shift.py} & §4.2 & standard vs formatting vs joint
ranking shift \\
\texttt{turn\_depth\_analysis.py} & §4.3 & formatting × vote-time
conversation-depth interactions \\
\texttt{topic\_analysis.py} & §4.4 & within-topic fits + topic × style
controls \\
\texttt{extract\_prompts.py} + \texttt{task\_classify.py} +
\texttt{task\_analysis.py} & §4.4 & task proxy and within-task fits \\
\texttt{robustness\_random.py}; \texttt{time\_block\_bootstrap.py} &
§4.5 & random-only coefficients; weekly block bootstrap \\
\texttt{mattr\_stress.py} + \texttt{mattr\_alt.py} +
\texttt{analyze\_mattr\_alt.py} & §4.5 & MATTR length sensitivity,
strata, and MTLD/content-word/no-proper-noun variants \\
\texttt{external\_leaderboard\_analysis.py} & §4.6 &
capability-benchmark and LMArena correlations, snapshot provenance, and
model-match audits \\
\texttt{generate\_external\_figure.py} & §4.6 & capability-benchmark
correlation differences and intervals \\
Live leaderboard audit & §4.6 & dated production examples in
\texttt{results/production\_ranking\_examples.json}, kept distinct from
the research release \\
\texttt{audit\_vote\_timing.py} & §2.3, §4.3 & raw-vote, final-turn,
visible-depth, and post-vote-gap validation \\
\texttt{audit\_reasoning\_content.py} & §2.3 & hidden-reasoning
serialisation and parser-recovery audit \\
\texttt{endogeneity\_analysis.py} & §5.2 & between-model composition and
tier heterogeneity \\
\texttt{qualitative\_analysis.py} & §5.2 & winner-flip prevalence and
asymmetry \\
\end{longtable}
}

\section{Complete External-Alignment
Intervals}\label{appendix-b-external-intervals}

All entries are the change in Spearman correlation relative to the raw
Compar:IA ranking, with paired 95\% bootstrap intervals from 10,000
matched-model resamples. These tables report the uncertainty for both
adjustments rather than only the formatting-controlled comparison shown
in Table 7B.

{\def\LTcaptype{none}
\footnotesize
\setlength{\tabcolsep}{4pt}
\begin{longtable}[]{@{}lcc@{}}
\toprule\noalign{}
Capability benchmark & Formatting-controlled − raw &
Full joint-controlled − raw \\
\midrule\noalign{}
\endhead
\bottomrule\noalign{}
\endlastfoot
Epoch Capabilities Index & −0.014 {[}−0.079, +0.054{]} &
−0.081 {[}−0.199, +0.005{]} \\
GPQA Diamond & −0.011 {[}−0.063, +0.035{]} &
−0.088 {[}−0.223, +0.011{]} \\
FrontierMath & −0.044 {[}−0.223, 0.000{]} &
−0.165 {[}−0.527, +0.035{]} \\
LiveBench & −0.142 {[}−0.485, +0.097{]} &
−0.061 {[}−0.412, +0.272{]} \\
ARC-AGI-2 & −0.049 {[}−0.367, +0.209{]} &
−0.235 {[}−0.806, +0.182{]} \\
SciCode & −0.014 {[}−0.149, +0.073{]} &
−0.077 {[}−0.364, +0.022{]} \\
Aider Polyglot & −0.036 {[}−0.314, +0.199{]} &
+0.158 {[}−0.192, +0.549{]} \\
\end{longtable}
}

{\def\LTcaptype{none}
\footnotesize
\setlength{\tabcolsep}{4pt}
\begin{longtable}[]{@{}lcc@{}}
\toprule\noalign{}
LMArena preference ranking & Formatting-controlled − raw &
Full joint-controlled − raw \\
\midrule\noalign{}
\endhead
\bottomrule\noalign{}
\endlastfoot
Raw, overall & +0.008 {[}−0.045, +0.066{]} &
−0.082 {[}−0.162, −0.010{]} \\
Style-controlled, overall & +0.040 {[}−0.018, +0.115{]} &
−0.033 {[}−0.120, +0.050{]} \\
Raw, French & +0.017 {[}−0.064, +0.109{]} &
−0.082 {[}−0.183, +0.022{]} \\
Style-controlled, French & +0.072 {[}−0.027, +0.190{]} &
−0.008 {[}−0.131, +0.130{]} \\
\end{longtable}
}

\end{document}